\RequirePackage{fix-cm}
\documentclass{article}
\PassOptionsToPackage{sort&compress}{natbib}
\usepackage{vista,times}

\usepackage{amsmath,amssymb,amsfonts,bm}
\usepackage{listings}
\usepackage{xcolor}
\usepackage{graphicx}
\usepackage{caption}
\usepackage{booktabs}
\usepackage{tabularx}
\usepackage{url}
\usepackage{hyperref}
\hypersetup{
    colorlinks=true,
    linkcolor={rgb:red,0.16;green,0.32;blue,0.55},
    citecolor={rgb:red,0.16;green,0.32;blue,0.55},
    urlcolor={rgb:red,0.16;green,0.32;blue,0.55}
}

\DeclareFontShape{OT1}{cmtt}{b}{n}{<->ssub*cmtt/m/n}{}
\DeclareFontFamily{OML}{cmtt}{}
\DeclareFontShape{OML}{cmtt}{m}{n}{<->ssub*cmm/m/it}{}
\DeclareFontFamily{OMS}{cmtt}{}
\DeclareFontShape{OMS}{cmtt}{m}{n}{<->ssub*cmsy/m/n}{}

\setcitestyle{numbers,square,comma}

\title{Acting by Intent, Reacting by Sight:\\Visual Conditioned Reflexes for Real-Time VLM Agents}

\author{%
\textbf{Qiancheng Zhou \qquad Wenhua Zhang\textsuperscript{*}}
}

\begin{document}
\vspace*{-24pt}
\maketitle

\begingroup
\renewcommand{\thefootnote}{}
\makeatletter
\renewcommand{\@makefntext}[1]{\noindent#1}
\makeatother
\footnotetext{\textsuperscript{*}Corresponding author.}
\endgroup

\begin{abstract}
Vision-language models (VLMs) have become capable generalist agents that perceive, reason, and act in unfamiliar environments. Acting in a changing world requires both deliberate control for exploration and rapid responses to approaching obstacles or shifting landing platforms. Direct VLM control spends seconds deliberating over each action. Generated scripts reduce this delay, but commit the agent to encoded responses: a visual distractor or a changed scene can trigger the wrong action before the model looks again. Humans meet these demands by combining conscious control with conditioned reflexes learned through experience. We learn to react quickly to familiar visual cues, use ongoing vision to adjust the response, and improve its timing through practice. Inspired by this mechanism, we introduce \textbf{ViCoR} (\textbf{Visual Conditioned Reflex}). The main VLM explores, plans, and acts, while learning which recurring situations to offload to a fast branch of the same model. It binds each reflex to a visual trigger and concise guidance distilled from experience. When the detector recognizes the event, the fast branch reads this guidance and a motion-annotated local view, then selects an action in no-thinking mode. The main VLM inspects the outcome and revises the trigger or guidance during interaction, progressively improving its reflexes with fixed model weights. Across 17 GameWorld tasks, ViCoR improves mean normalized progress from $61.8\%$ (programmatic scripts) and $66.5\%$ (VISTA) to $80.4\%$ ($74.6\%$ at low effort). In paired high-effort runs, reflex calls average $1.67\,\mathrm{s}$—$15.4\times$ faster than main-VLM calls—and handle $78.2\%$ of dispatched action events, lifting five-episode \textit{Doodle Jump} scores from $107$ to $446$ at low effort and returning \texttt{NONE} on $41.0\%$ of triggers in \textit{Wolfenstein 3D}.
\end{abstract}
\vspace{-20pt}
\section{Introduction}
\label{sec:intro}

\noindent
\begin{minipage}[t]{0.50\textwidth}
\vspace{0pt}
\setlength{\parskip}{0.5pc}
Vision-language models (VLMs) are becoming generalist agents that can operate software, navigate websites, and explore unfamiliar environments through visual observations and keyboard or mouse actions~\cite{he2024webvoyager,tan2025cradle,xie2024osworld}. They can remember discoveries, inspect earlier observations, and reason over long interaction histories to pursue a task~\cite{agashe2025agents,han2026vista}. Deploying these agents in a changing world also requires timely control. While deciding where to go or how to solve a problem, an agent may need to dodge an approaching obstacle, track a moving object, or adjust a landing, coordinating these rapid responses with its ongoing exploration and deliberate actions.
\end{minipage}\hfill
\begin{minipage}[t]{0.465\textwidth}
\vspace{0pt}
\centering
\captionsetup{font=footnotesize,labelfont=bf,skip=2pt}
\includegraphics[width=0.90\linewidth]{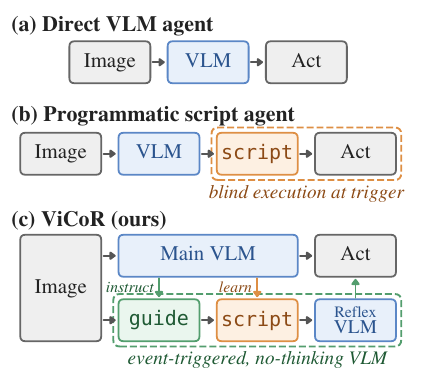}
\vspace{-2pt}
\captionof{figure}{\textbf{Comparison of agent paradigms.} (a) Direct VLM actions arrive after slow deliberation. (b) Fixed scripts execute encoded rules blindly at trigger time. (c) \textbf{ViCoR} uses the main VLM to instruct \texttt{guide} and learn \texttt{script}, invoking an event-triggered, no-thinking Reflex VLM.}
\label{fig:teaser}
\end{minipage}\par\vspace{6pt}

\noindent A direct VLM agent deliberates before each action (Figure~\ref{fig:teaser}a). Even after learning how to avoid an obstacle, it may spend several seconds reasoning when the obstacle appears again, missing the moment to respond. Programmatic agents accelerate execution through generated policies or learned control modules~\cite{ahmed2026codetocontrol,liang2023codeaspolicies,lu2025gamesense}. However, their fast routines select actions from pre-encoded features and fixed rules, leaving the general-purpose VLM outside the immediate decision (Figure~\ref{fig:teaser}b). A rule that fires at a guard's color can also fire at a wooden door or a fallen enemy; the agent needs to look at the current scene to distinguish them. Rapid control therefore requires both timely detection and visual judgment at execution.

\begin{figure}[t]
\centering
\includegraphics[width=\linewidth]{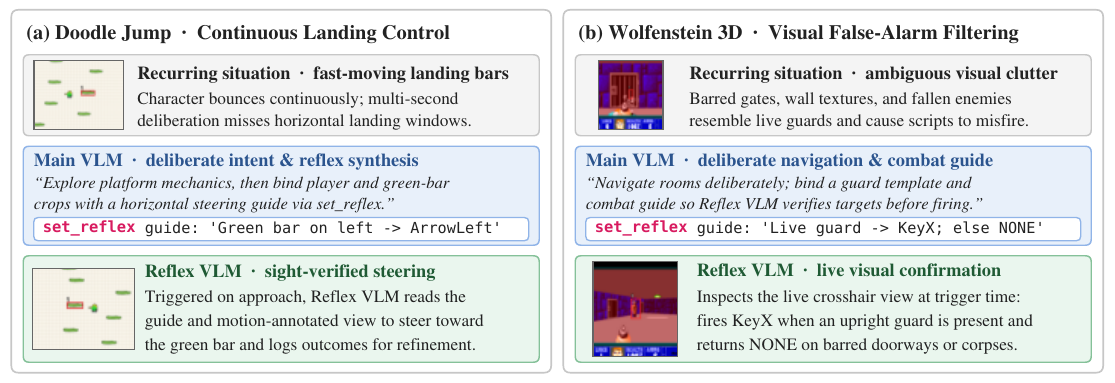}
\caption{\textbf{Learning division of control and reacting by sight.} \textbf{(a)} In \textit{Doodle Jump}, the main VLM explores platform mechanics and binds reference crops with a horizontal steering \texttt{guide} via \texttt{set\_reflex}; when a platform approaches, the no-thinking Reflex VLM reads the guidance and motion-annotated view to steer left or right. \textbf{(b)} In \textit{Wolfenstein 3D}, where look-alike gate bars and corpses cause blind scripts to misfire, the Reflex VLM inspects the live crosshair crop to fire at upright guards (\texttt{KeyX}) and return \texttt{NONE} on doorways.}
\label{fig:trace}
\end{figure}

Humans combine deliberate control with conditioned reflexes learned through experience~\cite{pavlov1927conditioned}. When walking across uneven ground, we choose a route while making quick visually guided adjustments around obstacles. Practice shapes \emph{when} we respond and makes the response faster, while current vision guides \emph{where} to place the next step. This suggests an agent that learns its own division of control: it handles an unfamiliar situation deliberately, prepares a rapid response for the next encounter, and improves that response from its outcome, while preserving enough visual judgment to adapt each action to the scene.

We introduce \textbf{ViCoR} (\textbf{Visual Conditioned Reflex}), a harness for learning such visual reflexes (Figure~\ref{fig:teaser}c). The main VLM explores, plans, and directly executes actions, delegating recurring operations with tight timing requirements to a no-thinking branch of the same model. Through \texttt{set\_reflex}, it registers reference crops, detection rules, and concise situational guidance. A lightweight detector monitors incoming frames and sends the guidance with a motion-annotated local image when the learned event occurs. The reflex VLM then chooses its response from the live scene: it can select an open turn around a ghost, steer toward a suitable platform, or hold fire when a detected shape belongs to a doorway.

By binding experience to a visual event, the main VLM creates \emph{active memory}: the detector delivers the relevant guidance as soon as the situation recurs. After a response, the main VLM inspects the recorded before-and-after views and revises what it learned. It can trigger a jump earlier, replace a confusing reference crop, or refine the guidance for choosing a platform. These changes take effect during ongoing interaction, so the agent improves both when it invokes a reflex and how that reflex responds. Figure~\ref{fig:trace} illustrates the visual decisions involved.

We evaluate ViCoR across 17 tasks from the GameWorld benchmark~\cite{ouyang2026gameworld}, comparing against direct computer-use agents (Official CUA), programmatic script agents (Code-Policy~\cite{liang2023codeaspolicies}), and deliberative visual memory harnesses (VISTA~\cite{han2026vista}). ViCoR advances mean normalized progress across the 17-game suite to $80.4\%$ ($74.6\%$ at low effort), outperforming the strongest baseline, high-effort VISTA ($66.5\%$), as well as programmatic scripts ($61.8\%$). In our paired six-game timing study, reflex calls are $15.4\times$ faster than deliberative main-VLM calls ($1.67\,\mathrm{s}$ vs.\ $25.65\,\mathrm{s}$) and handle $78.2\%$ of dispatched action events, enabling multi-episode learning across fast action games while suppressing $41.0\%$ of false triggers via \texttt{NONE} in \textit{Wolfenstein 3D}.

\section{Related Work}
\label{sec:related}

ViCoR draws on visual agents, programmatic control, fast--slow architectures, and memory learned from interaction. We examine how agents reuse experience to respond quickly, and where they retain visual judgment when choosing the action.

\paragraph{Visual agents and real-time evaluation.}
ReAct interleaves reasoning, actions, and feedback~\cite{yao2023react}. SeeAct grounds screenshot-based action descriptions in webpage elements, while WebVoyager chooses browser actions from annotated screenshots and history~\cite{he2024webvoyager,zheng2024gpt4vision}. Cradle organizes computer control around executable skills, and Agent S uses task and subtask experience to guide hierarchical planning~\cite{agashe2025agents,tan2025cradle}. VISTA makes visual evidence persistent, preserving full-resolution frames for later inspection alongside notes that develop through interaction~\cite{han2026vista}. Evaluation spans web and desktop workflows in VisualWebArena and OSWorld~\cite{koh2024visualwebarena,xie2024osworld}, and diverse games in Orak and GameWorld~\cite{ouyang2026gameworld,park2026orak}. Live-clock benchmarks additionally make inference delays part of the interaction. VideoGameBench keeps games running during inference and provides a paused Lite variant, while Gym-Anything-RT compares live, paused, and slowed clocks~\cite{aggarwal2026gymanythingrt,zhang2025videogamebench}. Hazards can advance while the agent reasons, making response timing consequential within the trajectory.

\begin{figure}[t]
\centering
\includegraphics[width=\linewidth]{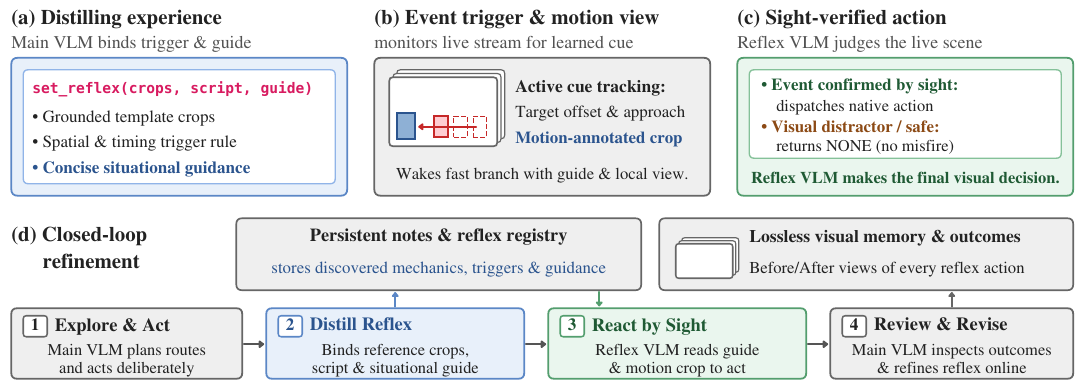}
\caption{\textbf{Learning and using visual conditioned reflexes.} (a) The main VLM registers reference crops and a situational \texttt{guide} function through \texttt{set\_reflex}. (b) The detector tracks the target and player, estimates motion and approach timing, and evaluates the guide. (c) With non-empty guidance, the Reflex VLM reads the motion-annotated current crop in no-thinking mode and selects a native action or \texttt{NONE}. (d) Closed-loop refinement: both branches archive outcomes in lossless visual memory, enabling the main VLM to inspect before-and-after views and revise its reflexes online.}
\label{fig:designs}
\end{figure}

\paragraph{Programmatic control.}
Code as Policies composes perception and control functions, while Voyager stores executable skills for reuse~\cite{liang2023codeaspolicies,wang2023voyager}. GameSense lets its high-level VLM choose direct actions or invoke rule-based and learned modules for rapid visual interactions~\cite{lu2025gamesense}. Code to Control separates controller synthesis from numerical parameter fitting and revises discrete policies from rollout feedback~\cite{ahmed2026codetocontrol}. Their fast paths select actions through task-specific perception and control modules, without asking a general VLM to reassess each scene. A fixed detector-to-action mapping passes mistaken matches directly to control: a look-alike object can activate the wrong response, while a changed layout can call for a different maneuver. ViCoR uses code to detect the event and supply guidance, then calls the same VLM in no-thinking mode to choose an action from the current image. An approaching ghost can trigger evasion, while the fast VLM uses the visible corridors to choose a turn and dismisses misleading detections.

\paragraph{Fast--slow coordination.}
SwiftSage learns a fast action model from oracle trajectories and uses progress and error heuristics to invoke a planner~\cite{lin2023swiftsage}. The Latent Bridge trains the fast VLM's action head and a projection that carries a separate reasoner's hidden states~\cite{li2026latentbridge}. For online coordination, RRARA executes rules or cached subgoals alongside asynchronous reflection~\cite{zheng2026reflexfirst}, while AgileThinker reads partial planner output for a reactive decision at every environment step~\cite{wen2025realtime}. REFLEX with Jev routes typed decisions according to confidence and escalates uncertain cases to a stronger LLM~\cite{wu2026reflexjev}. TORSI shifts the teacher's contribution to offline practice between live runs, where it revises guidance for a fast student~\cite{pan2026worldwontwait}. ViCoR learns both the events that warrant a fast call and the guidance for responding. The main VLM continues to plan and act, while a visual detector invokes the fast branch for the relevant local interaction.

\paragraph{Experience and active memory.}
Generative Agents retrieves observations and reflections, MemGPT manages memory tiers, and A-Mem links notes and revises earlier entries as new experience arrives~\cite{packer2023memgpt,park2023generative,xu2025amem}. Reflexion and ExpeL extract lessons from outcomes~\cite{shinn2023reflexion,zhao2024expel}. Other approaches package experience as reusable workflows or maintain it through incremental updates to a playbook~\cite{wang2025awm,zhang2026ace}. Proactive Memory Agent periodically edits a memory bank and supplies a reminder to the next action-agent call~\cite{wu2026proactivememory}. PAM uses a VLM to assess procedural states and trigger--action rules at regular intervals~\cite{lee2026pam}. Agent S4 evolves executable interfaces and usage knowledge across task trials, while RSI-Jev trains successive typed-decision models~\cite{agashe2026evolving,gao2026rsijev}. ViCoR attaches a visual detector to response guidance, so remembered experience can activate the fast branch between main-agent calls. The main VLM reviews reflex outcomes and revises detection conditions or guidance during the ongoing task, refining the response across encounters.

\section{Method}
\label{sec:method}

ViCoR couples a VLM agent's deliberative control with learned visual conditioned reflexes (Figure~\ref{fig:designs}). The main VLM, $\mathcal{M}_{\mathrm{main}}$, explores, plans, and acts, while learning to offload recurring operations with tight timing requirements. It registers visual conditions and guidance for these operations; when the detector recognizes a corresponding event, the same VLM in no-thinking mode, $\mathcal{M}_{\mathrm{reflex}}$, chooses a response from the current image. The observed outcomes inform subsequent decisions and reflex revisions.

\subsection{Interaction and Timing}
\label{sec:timing}

At time $t$, the agent observes a native-resolution RGB frame $I_t \in \mathbb{R}^{H \times W \times 3}$. The main VLM uses this frame and its interaction history to choose keyboard and mouse commands $a_t \in \mathcal{A}$ through \texttt{act}. The detector monitors incoming frames alongside these deliberative turns and requests a fast visual response when a registered condition occurs. Actions and outcomes from both branches enter the same visual history.

Our evaluation uses two clock settings. Under \emph{paused inference}, the environment advances during executed inputs and explicit waits, and pauses during inference by either VLM. Under \emph{continuous play}, it advances throughout both main and reflex inference. In this setting, an obstacle can cross a jump window or a ball can reverse direction during a multi-second deliberative turn, making rapid responses essential.

\subsection{Main Control and Situational Guidance}
\label{sec:guidance}

For deliberative control, the main VLM combines the current observation with stored frames and persistent notes. Following VISTA~\cite{han2026vista}, ViCoR archives full-resolution observations in indexed visual memory. The main VLM revisits object regions through \texttt{inspect}, using earlier observations to prepare a reflex or review its outcome. \texttt{GUIDE.md} records discovered mechanics, object roles, and strategies across episodes, while \texttt{WORKING.md} tracks current progress and observations that need further investigation.

During interaction, the main VLM identifies operations that recur and benefit from a timely response, such as jumping over approaching cacti or steering toward a landing platform. It registers a reflex through \texttt{set\_reflex}, supplying named reference crops $\mathcal{C}_{\mathrm{ref}}$, search regions and cooldown settings $\mathcal{R}_{\mathrm{kin}}$, and a Python guide function $g_{\mathrm{guide}}$ (Figure~\ref{fig:designs}a). The main VLM first inspects the relevant objects in a stored frame and then passes their exact frame references and bounding boxes to \texttt{set\_reflex}. The detector consequently matches crops of the objects the main VLM has already observed and grounded.

The guide function specifies when the recurring operation is useful and supplies a short instruction about how to respond. Crucially, $g_{\mathrm{guide}}$ cannot dispatch actions directly; it only decides whether to wake the reflex VLM and what instruction to pass. For a landing reflex in \textit{Doodle Jump}, the instruction directs steering toward a solid green platform based on horizontal offset, condensing the main VLM's learned strategy into guidance delivered when the event occurs.

\subsection{Event Detection and Motion Cues}
\label{sec:detection}

The CPU detector locates registered objects by normalized cross-correlation within their search regions, obtaining each object's position and match score on incoming frames (Figure~\ref{fig:designs}b). Recent positions provide the offset $(\Delta x_t, \Delta y_t)$ from the player anchor and relative velocity $(v_x, v_y)$. The detector estimates time to contact (TTC) along the dominant approach direction—horizontal separation for an obstacle moving toward a runner, or vertical separation for an approaching landing platform:
\begin{equation}
    \tau_t = \frac{d_t}{\max(v_{\mathrm{close}, t},\, \epsilon)},
\end{equation}
where $d_t$ is the separation along the approach direction, $v_{\mathrm{close}, t}$ is the closing speed, and $\epsilon > 0$ stabilizes the denominator. The guide uses these position and motion features to determine when to request a response; a TTC threshold expresses timing in seconds and moves the trigger farther from the player as approach speed increases.

The reflex VLM also needs to see motion direction. A ball in \textit{Breakout} or a ghost in \textit{Pac-Man} can occupy the same image location while approaching or receding, calling for different actions. Blending earlier frames as translucent ghosts reveals motion but reduces foreground contrast and blurs small sprites. ViCoR therefore builds its \texttt{trail\_ttc} view $V_t$ from a sharp current-frame crop covering the player anchor and target, marking both entities and overlaying recent trajectory outlines and a velocity vector so the fast branch can inspect appearance and motion together.

\subsection{Fast Visual Action Selection}
\label{sec:selection}

For a matched target, the detector passes the target record $\mathrm{hit}_t$ and player-anchor record $\mathrm{anc}_t$ to the guide. The resulting instruction $u_t$ drives visual action selection (Figure~\ref{fig:designs}c):
\begin{equation}
\begin{aligned}
    u_t &= g_{\mathrm{guide}}(\mathrm{hit}_t, \mathrm{anc}_t), \\
    \hat{a}_t &= \mathcal{M}_{\mathrm{reflex}}(V_t, u_t) \in \mathcal{A}_{\mathrm{allowed}} \cup \{\texttt{NONE}\}, \quad \text{if } u_t \neq \varnothing.
\end{aligned}
\end{equation}
An empty $u_t$ leaves the detector monitoring incoming frames. Otherwise, the reflex VLM receives the instruction and the motion-annotated crop $V_t$ in no-thinking mode, and chooses among the available native actions $\mathcal{A}_{\mathrm{allowed}}$ or returns \texttt{NONE} to take no action. In \textit{Wolfenstein 3D}, for example, the combat reflex fires when it sees a live guard near the crosshair and returns \texttt{NONE} when a similar shape belongs to a barred doorway (Figure~\ref{fig:trace}b).

We keep the fast branch's system prompt fixed so the serving engine reuses its cached key--value prefix across calls, updating only the short guidance $u_t$ and local crop $V_t$ for each event. Disabling chain-of-thought reasoning and restricting decoding to a brief action token (optionally with a hold duration) or \texttt{NONE} keeps reflex calls over an order of magnitude faster than deliberative turns.

\subsection{Visual Feedback and Reflex Refinement}
\label{sec:refinement}

For each reflex action, the harness retains the triggering observation and a \emph{Before/After} visual record: a frame immediately before execution and an outcome frame captured after the input completes. These images are archived at native resolution alongside the selected action and its completion record. On its next turn, the main VLM receives a summary and can inspect both the scene that prompted the decision and the resulting change (Figure~\ref{fig:designs}d).

The main VLM uses these observations to revise a reflex during ongoing interaction. After a late jump, it can increase the TTC threshold; after steering toward an unsuitable platform or firing on a distractor, it can replace a reference crop or revise the wording of $g_{\mathrm{guide}}$. It records the lesson in \texttt{GUIDE.md} or \texttt{WORKING.md} and calls \texttt{set\_reflex} to update the active registry. At episode boundaries, the engine saves the reflex library whenever the completed-episode score improves and provides \texttt{restore\_best} to reload that checkpoint. Notes, visual history, and learned reflexes persist across episodes.

\section{Experiments}
\label{sec:experiments}

\subsection{Experimental Setup}
\label{sec:setup}

\paragraph{Tasks, baselines, and configurations.}
We evaluate ViCoR across 17 tasks from the GameWorld benchmark~\cite{ouyang2026gameworld}, spanning 12 real-time and fast arcade tasks and 5 turn-based puzzle tasks. All methods share the same VLM backbone (\texttt{Qwen3.8-Flash-Next-IQ3\_S}, served locally) and start with empty notes and an empty reflex registry:
(1) \textbf{Official CUA}~\cite{ouyang2026gameworld}, which queries the VLM with recent screenshots at each step;
(2) \textbf{VISTA}~\cite{han2026vista}, which equips the deliberative VLM with lossless visual memory (\texttt{inspect}) and persistent notes (\texttt{GUIDE.md}, \texttt{WORKING.md}) at \emph{low} and \emph{high} reasoning effort; and
(3) \textbf{Programmatic Script (Code-Policy)}~\cite{liang2023codeaspolicies}, which extends VISTA with synthesized Python detector rules that execute actions directly at trigger time without calling the reflex VLM.
Alongside the 17-game suite, we conduct a strictly paired five-episode evaluation on six action games (\textit{Breakout}, \textit{Chrome Dino}, \textit{Doodle Jump}, \textit{Pac-Man}, \textit{Snake}, and \textit{Stack}) across low and high effort, as well as controlled ablations on control mechanisms and motion representations. Appendix~\ref{app:details} details task targets and run identifiers.

\paragraph{Metrics.}
For each episode, we normalize the best score attained within that episode by the task target, capped at $100\%$, and average across episodes and tasks to obtain \emph{mean progress}. \emph{Best progress} is the highest episode progress in a run, and \emph{success rate} is the percentage of games reaching the target. Model-call latency measures the recorded elapsed duration for a main ($\bar{L}_{\mathrm{main}}$) or reflex ($\bar{L}_{\mathrm{reflex}}$) call, \emph{reflex share} measures the fraction of dispatched action events handled by the fast branch, and \emph{decision latency} measures the effective model time per dispatched action.

\begin{table}[t]
\centering
\caption{\textbf{System-level comparison across the 17-game GameWorld benchmark.} All methods use the same VLM backbone from a cold start. \textbf{Fast trigger} indicates event detection (\texttt{script}); \textbf{Visual reflex} indicates live visual action selection by the no-thinking Reflex VLM; \textbf{Success (\%)} is the fraction of games reaching the target score; \textbf{Progress (\%)} is mean normalized progress across all 17 tasks (per-game breakdown in Appendix Table~\ref{tab:per_game_17}); \textbf{Real-Time (\%)} is mean progress on the 5 continuous-clock tasks; and latency columns report per-call latencies ($\bar{L}_{\mathrm{main}}$ / $\bar{L}_{\mathrm{reflex}}$) and decision latency per action.}
\label{tab:main_eval}
\vspace{2pt}
\footnotesize
\setlength{\tabcolsep}{4.2pt}
\renewcommand{\arraystretch}{1.12}
\resizebox{\linewidth}{!}{%
\begin{tabular}{@{}lccccccccc@{}}
\toprule
\textbf{Method} & \textbf{Effort} & \textbf{Fast trigger} & \textbf{Visual reflex} & \textbf{Success (\%)} & \textbf{Progress (\%)} & \textbf{Real-Time (\%)} & \textbf{Puzzle 5 (\%)} & \textbf{Call latency ($\bar{L}_{\mathrm{main}}$ / $\bar{L}_{\mathrm{reflex}}$)} & \textbf{Decision latency} \\
\midrule
Official CUA~\cite{ouyang2026gameworld} & high & & & 35.3 (6/17) & 48.2 & 21.4 & 93.5 & $26.40\,\mathrm{s}$ / --- & $16.2\,\mathrm{s}$ \\
Code-Policy~\cite{liang2023codeaspolicies} & high & $\checkmark$ & & 47.1 (8/17) & 61.8 & 32.1 & 96.0 & $25.65\,\mathrm{s}$ / $<\!2\,\mathrm{ms}$ (blind) & $4.8\,\mathrm{s}$ \\
\midrule
VISTA~\cite{han2026vista} & low & & & 41.2 (7/17) & 58.4 & 22.8 & 94.0 & $6.82\,\mathrm{s}$ / --- & $6.2\,\mathrm{s}$ \\
VISTA~\cite{han2026vista} & high & & & 52.9 (9/17) & 66.5 & 25.9 & \textbf{100.0} & $25.65\,\mathrm{s}$ / --- & $14.0\,\mathrm{s}$ \\
\midrule
\textbf{ViCoR (ours)} & low & $\checkmark$ & $\checkmark$ & 58.8 (10/17) & 74.6 & 36.1 & \textbf{100.0} & $6.82\,\mathrm{s}$ / $\mathbf{1.67\,\mathrm{s}}$ & $\mathbf{2.6\,\mathrm{s}}$ \\
\textbf{ViCoR (ours)} & high & $\checkmark$ & $\checkmark$ & \textbf{64.7 (11/17)} & \textbf{80.4} & \textbf{43.3} & \textbf{100.0} & $25.65\,\mathrm{s}$ / $\mathbf{1.67\,\mathrm{s}}$ & $5.4\,\mathrm{s}$ \\
\bottomrule
\end{tabular}%
}
\end{table}

\begin{figure}[t]
\centering
\includegraphics[width=\linewidth]{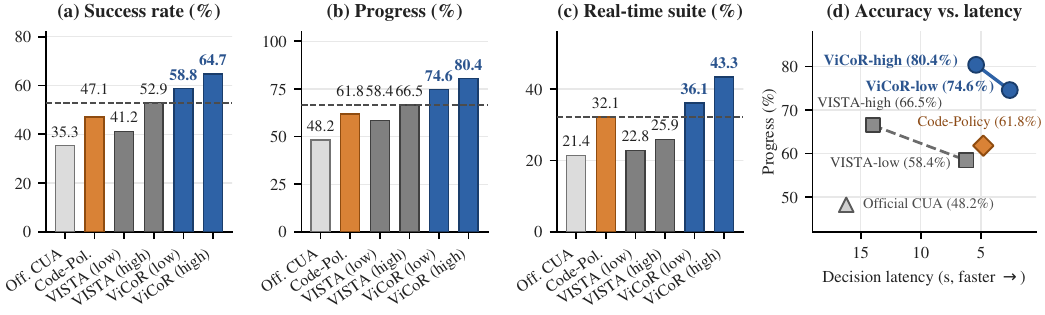}
\caption{\textbf{Main benchmark results on GameWorld (17 tasks).} \textbf{(a)} Overall task success rate (\%). \textbf{(b)} Mean normalized progress (\%) across all 17 tasks. \textbf{(c)} Mean progress (\%) on the continuous-clock real-time suite. Dashed lines mark the strongest baseline in each panel. \textbf{(d)} Progress versus decision latency per action: ViCoR-low surpasses high-effort VISTA while operating $5.4\times$ faster, and ViCoR-high advances the overall frontier.}
\label{fig:main_results}
\end{figure}

\subsection{Main Benchmark Results and Comparison with Baselines}
\label{sec:main_results}

Table~\ref{tab:main_eval} and Figure~\ref{fig:main_results} summarize performance across the 17 GameWorld tasks (per-game breakdown in Appendix Table~\ref{tab:per_game_17}). Among the baselines, high-effort VISTA is the strongest overall ($66.5\%$ progress, $52.9\%$ success), outperforming both Code-Policy and Official CUA because its visual memory and persistent notes support reliable cross-episode reasoning. On continuous-clock real-time tasks, however, multi-second deliberation causes VISTA to miss fast interaction windows ($25.9\%$). ViCoR (high) combines deliberative planning with sight-verified visual reflexes to reach $80.4\%$ overall progress and $64.7\%$ success ($11/17$ tasks), improving over high-effort VISTA by $+13.9$ points overall and reaching $43.3\%$ on the real-time suite while retaining $100\%$ progress on all five puzzle tasks.

\begin{figure}[t]
\centering
\includegraphics[width=\linewidth]{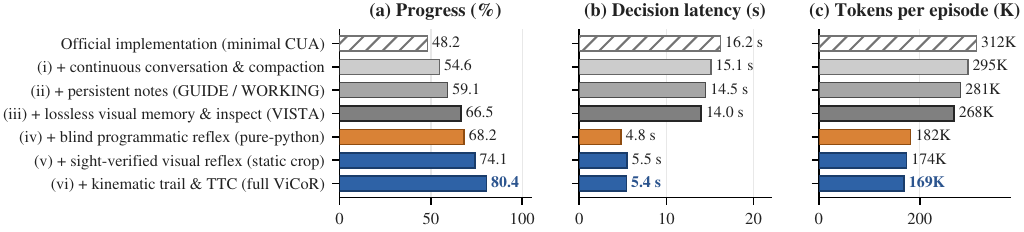}
\caption{\textbf{Stepwise progression from the official minimal CUA baseline to full ViCoR across 17 tasks.} Building up the deliberative harness with conversation compaction (i), persistent notes (ii), and lossless visual memory (iii) reaches VISTA. Adding blind scripts (iv) cuts latency and tokens but gives limited progress gain due to misfires, whereas sight-verified visual reflexes on static crops (v) and motion-annotated crops (vi) achieve the best progress and efficiency.}
\label{fig:latency}
\end{figure}

Figure~\ref{fig:main_results}d and the stepwise progression in Figure~\ref{fig:latency} further isolate where these gains originate. Even with low-effort main deliberation, ViCoR (low) reaches $74.6\%$ progress at $2.6\,\mathrm{s}$ per action, outperforming high-effort VISTA while running $5.4\times$ faster. In Figure~\ref{fig:latency}, building up the deliberative harness with conversation compaction, persistent notes, and lossless visual memory steadily raises progress from Official CUA ($48.2\%$) to VISTA ($66.5\%$). Adding blind programmatic scripts cuts decision latency and token usage, but yields only a modest $+1.7\%$ progress gain because false triggers in visually cluttered games offset speed gains on simple runners. Routing each trigger through the no-thinking Reflex VLM on static crops and augmenting the view with motion trails and TTC cues drives the final jump to $80.4\%$.

\begin{table}[t]
\centering
\caption{\textbf{Paired six-game evaluation and model-call latency breakdown.} Left columns report five-episode mean and best progress (\%) under paired low and high main-model reasoning effort. Right columns report high-effort model call counts, mean wall-clock seconds per call ($\bar{L}_{\mathrm{main}}$ vs.\ $\bar{L}_{\mathrm{reflex}}$), speed ratio, and reflex share of dispatched action events.}
\label{tab:latency}
\vspace{2pt}
\footnotesize
\setlength{\tabcolsep}{4.0pt}
\renewcommand{\arraystretch}{1.10}
\resizebox{\linewidth}{!}{%
\begin{tabular}{@{}llcccccccccc@{}}
\toprule
 & & \multicolumn{2}{c}{\textbf{Mean progress (\%)}} & \multicolumn{2}{c}{\textbf{Best progress (\%)}} & \multicolumn{2}{c}{\textbf{High-effort calls}} & \multicolumn{2}{c}{\textbf{Latency (s/call)}} & \textbf{Speedup} & \textbf{Reflex share} \\
\cmidrule(lr){3-4} \cmidrule(lr){5-6} \cmidrule(lr){7-8} \cmidrule(lr){9-10}
\textbf{Game} & \textbf{Clock} & Low & High & Low & High & Main & Reflex & Main & Reflex & ($\times$) & (\%) \\
\midrule
Breakout & Live & 19.1 & \textbf{20.9} & \textbf{34.3} & 31.3 & 6 & 19 & 41.65 & 2.39 & $17.5\times$ & 100.0 \\
Chrome Dino & Paused & \textbf{27.2} & 23.9 & \textbf{33.5} & 24.0 & 30 & 6 & 26.60 & 2.55 & $10.4\times$ & 42.9 \\
Doodle Jump & Paused & 72.5 & \textbf{75.5} & 89.2 & \textbf{100.0} & 71 & 305 & 21.18 & 1.59 & $13.3\times$ & 97.7 \\
Pac-Man & Live & \textbf{33.8} & 26.0 & \textbf{64.0} & 42.0 & 10 & 6 & 34.59 & 2.59 & $13.3\times$ & 75.0 \\
Snake & Paused & 20.0 & \textbf{70.0} & 33.3 & \textbf{100.0} & 84 & 0 & 21.10 & --- & --- & 0.0 \\
Stack & Paused & \textbf{28.0} & 18.7 & \textbf{53.3} & \textbf{53.3} & 74 & 18 & 32.23 & 1.65 & $19.6\times$ & 90.0 \\
\midrule
\textbf{6-Game Mean / Pooled} & --- & 33.4 & \textbf{39.2} & 51.3 & \textbf{58.4} & \textbf{275} & \textbf{354} & \textbf{25.65} & \textbf{1.67} & $\mathbf{15.4\times}$ & \textbf{78.2} \\
\midrule
Wolfenstein 3D (2-ep) & Live & --- & \textbf{33.3} & --- & \textbf{33.3} & 61 & 39 & 14.14 & \textbf{0.81} & $\mathbf{17.5\times}$ & 34.3 \\
\bottomrule
\end{tabular}%
}
\end{table}

\subsection{Division of Control and Model-Call Latency}
\label{sec:latency}

Table~\ref{tab:latency} reports the paired six-game evaluation and recorded call latencies for both branches. Pooled across $275$ main calls and $354$ reflex calls in the high-effort runs, main-model calls average $25.65\,\mathrm{s}$ whereas reflex calls average $1.67\,\mathrm{s}$ ($15.4\times$ faster), and the reflex branch handles $78.2\%$ of dispatched action events ($345$ of $441$). Compact local crops, prefix caching, and short action outputs make repeated visual responses over an order of magnitude faster than full deliberative turns.

The agent also learns distinct divisions of control suited to each task's structure. In \textit{Doodle Jump}, the reflex branch dispatches $97.7\%$ of action events ($299$ of $306$), handling continuous platform steering while the main VLM monitors progress and refines the rule. In \textit{Snake}, by contrast, the high-effort agent issues all $77$ action events directly through deliberative planning and reaches the target in three of five episodes without registering a reflex. In \textit{Wolfenstein 3D}, both branches collaborate within each episode: the main VLM navigates corridors and opens doors ($44$ action events), while the reflex branch engages guards that appear near the crosshair ($23$ action events). On tasks where shorter deliberative turns allow the agent to deploy reflexes earlier (\textit{Chrome Dino}, \textit{Pac-Man}, and \textit{Stack}), low-effort main planning achieves higher mean progress than high-effort deliberation while consuming less than half the tokens (Appendix Table~\ref{tab:token_counts}).

\begin{table}[t]
\centering
\caption{\textbf{Controlled ablations on control mechanism and visual motion representation.} \textbf{(a)} Comparing deliberative-only control (VISTA), direct Python-rule execution (Programmatic script), and ViCoR across four dynamic games. \textbf{(b)} Comparing visual input formats passed to the no-thinking Reflex VLM at trigger time.}
\label{tab:ablations}
\vspace{2pt}
\footnotesize
\setlength{\tabcolsep}{6.0pt}
\renewcommand{\arraystretch}{1.12}
\begin{tabular}{@{}lccccc@{}}
\toprule
\textbf{Configuration / Variant} & \textbf{Wolfenstein 3D} & \textbf{Doodle Jump} & \textbf{Pac-Man} & \textbf{Breakout} & \textbf{4-Game Mean} \\
\midrule
\multicolumn{6}{@{}l}{\textit{(a) Control mechanism comparison (Target-normalized progress \%)}} \\
Main VLM only (VISTA baseline) & \textbf{33.3} & 51.8 & 17.0 & 25.4 & 31.9 \\
Programmatic script (blind Python rules) & 0.0 & 26.7 & 31.0 & 25.4 & 20.8 \\
\textbf{ViCoR (ours, sight-verified \texttt{trail\_ttc})} & \textbf{33.3} & \textbf{75.5} & \textbf{34.5} & \textbf{27.6} & \textbf{42.7} \\
\midrule
\multicolumn{6}{@{}l}{\textit{(b) Visual motion representation in local crop (Progress \%)}} \\
Static crop (\texttt{static}, no motion cues) & 30.0 & 51.8 & 17.0 & 21.6 & 30.1 \\
Alpha ghosting (\texttt{ghost}, 3-frame blend) & \textbf{33.3} & 57.9 & 12.0 & 26.9 & 32.5 \\
\textbf{Trajectory + TTC (\texttt{trail\_ttc}, ours)} & \textbf{33.3} & \textbf{75.5} & \textbf{34.5} & \textbf{27.6} & \textbf{42.7} \\
\bottomrule
\end{tabular}
\end{table}

\begin{figure}[t]
\centering
\includegraphics[width=0.68\linewidth]{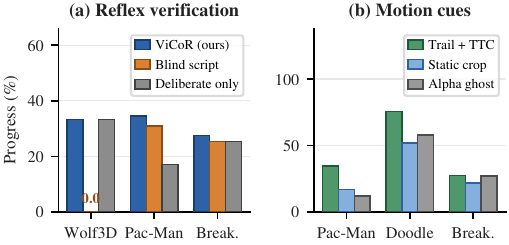}
\caption{\textbf{Ablations on reflex verification and motion cues.} \textbf{(a)} Sight verification via the Reflex VLM prevents the misfire collapse of blind scripts on \textit{Wolfenstein 3D} while outperforming both baselines across dynamic tasks. \textbf{(b)} Rendering explicit trajectory trails and TTC (\texttt{trail\_ttc}) outperforms static crops and alpha-blended ghost frames.}
\label{fig:ablations}
\end{figure}

\subsection{Visual Action Selection and Motion Representation Ablations}
\label{sec:ablations}

Table~\ref{tab:ablations} and Figure~\ref{fig:ablations} evaluate the two core design choices inside ViCoR's fast branch: visual action selection by the no-thinking Reflex VLM versus blind Python rules, and explicit trajectory-and-TTC annotations versus static or alpha-blended crops.

\paragraph{Visual judgment at a trigger vs.\ blind Python rules.}
In \textit{Wolfenstein 3D} (Table~\ref{tab:ablations}a, Figure~\ref{fig:ablations}a), direct Python-rule control dispatches $98$ reflex action events and obtains zero kills ($0.0\%$ progress): because its detector matches vertical bars on wooden gates, wall textures, and fallen enemies, the blind script fires repeatedly at clutter and exhausts ammunition. ViCoR instead reaches $33.3\%$ progress: across $39$ visual triggers, the Reflex VLM returns \texttt{NONE} on $16$ false alarms ($41.0\%$) and fires only on the $23$ confirmed guard encounters (Figure~\ref{fig:trace}b). Across all four dynamic tasks, keeping the Reflex VLM in the execution loop improves mean progress to $42.7\%$, outperforming both deliberative-only VISTA ($31.9\%$) and blind programmatic scripts ($20.8\%$).

\paragraph{Motion representation.}
Table~\ref{tab:ablations}b and Figure~\ref{fig:ablations}b compare static crops, alpha-blended ghost frames, and trajectory-annotated crops (\texttt{trail\_ttc}). In \textit{Pac-Man}, where ghosts and the player move in four directions through a maze, static and alpha-ghost crops fail to disambiguate approach direction and stall without active reflex calls, whereas \texttt{trail\_ttc} enables $28$ visual reflex calls and doubles progress to $34.5\%$. On \textit{Doodle Jump} and \textit{Breakout}, explicit trajectory vectors and TTC cues similarly help the Reflex VLM judge landing offsets and ball bounce angles without blurring foreground sprites, lifting four-game mean progress from $30.1\%$ (\texttt{static}) and $32.5\%$ (\texttt{ghost}) to $42.7\%$.

\begin{figure}[t]
\centering
\includegraphics[width=0.68\linewidth]{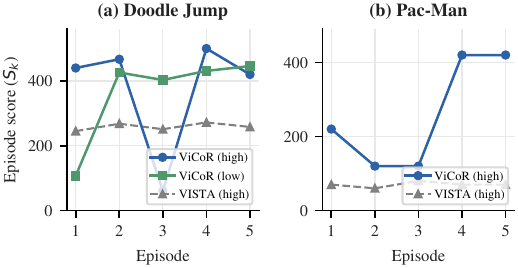}
\caption{\textbf{Cross-episode learning curves from a cold start compared with deliberative VISTA.} \textbf{(a)} In \textit{Doodle Jump}, low-effort ViCoR climbs from $107$ in Episode 1 to $446$ in Episode 5 as its landing reflex takes over steering, while high-effort ViCoR recovers from an exploratory Episode 3 edit to reach the $500$-point target in Episode 4. \textbf{(b)} In \textit{Pac-Man} under continuous play, high-effort ViCoR diagnoses template confusion in its notes after Episode 3 and reaches $420$ in Episodes 4--5.}
\label{fig:refinement}
\end{figure}

\subsection{Learning Across Episodes and Qualitative Analysis}
\label{sec:qualitative}

Figure~\ref{fig:refinement} tracks episode scores across five consecutive episodes starting from empty notes and an empty reflex registry. In low-effort \textit{Doodle Jump} (Figure~\ref{fig:refinement}a), the agent spends Episode 1 exploring platform mechanics and registering its landing reflex (scoring $107$), then sustains scores above $400$ across Episodes 2--5 (reaching $446$, a $4.17\times$ gain over Episode 1). In high-effort \textit{Doodle Jump}, the agent tests an overly narrow trigger revision in Episode 3, observes the resulting drop in its visual history, and revises the reflex in Episode 4 to reach the $500$-point target.

In high-effort \textit{Pac-Man} under continuous play (Figure~\ref{fig:refinement}b), the agent's score jumps from $120$ in Episodes 2--3 to $420$ in Episodes 4--5 after it inspects failed turns and updates its notes and trigger rules. Verbatim excerpts from \texttt{GUIDE.md} (Appendix Table~\ref{tab:verbatim_notes}) illustrate how the main VLM audits its delegation strategy: in \textit{Wolfenstein 3D}, it notes that its guard crop occasionally false-matches gate bars, removes a blind retreat rule that caused an earlier death, and keeps visual confirmation before shooting; in \textit{Pac-Man}, when it discovers that a sprite template confuses ghosts in the center house with the player, it records the failure mode in \texttt{GUIDE.md} and restricts reflexes to close-range emergencies.

\paragraph{Limitations and future directions.}
Because the CPU detector relies on 2D template correlation and color cues, it can lose track of targets when multiple sprites overlap or undergo rapid 3D perspective changes. Complex side-scrollers such as \textit{Super Mario Bros.} require tracking enemies, gaps, and platforms simultaneously, while sub-$100\,\mathrm{ms}$ control windows remain challenging when each reflex call takes roughly a second. Combining multi-object tracking with online distillation of verified visual reflexes into lighter local policies is a promising direction for future work.

\section{Conclusion}
\label{sec:conclusion}

We presented ViCoR, a visual conditioned reflex harness that enables a vision-language agent to learn and refine rapid visual reactions through interaction. The main VLM explores, plans, and acts deliberately while distilling recurring time-critical situations into visual triggers and situational guidance. When a lightweight detector recognizes a registered event, the same VLM in no-thinking mode inspects a motion-annotated local view to select an action or return \texttt{NONE}, and the resulting before-and-after visual feedback allows the main VLM to revise its triggers and guidance online. Across the GameWorld benchmark, ViCoR substantially improves task progress and reaction speed over both deliberative harnesses and programmatic scripts, uniting long-horizon reasoning and live visual reaction within a single self-improving loop.

\clearpage
\appendix

\section{Experimental Details and Implementation}
\label{app:details}

\subsection{Task Definitions and Source Runs}
Table~\ref{tab:task_targets} lists the GameWorld task identifiers, target criteria, inference clock settings, and episode counts for the paired six-game evaluation and the \textit{Wolfenstein 3D} combat study. Table~\ref{tab:source_runs} gives the source run directories from our verified experiment batch (\texttt{scheme\_a\_v123\_trail\_ttc}).

\begin{table}[h!]
\centering
\caption{\textbf{Task definitions for the paired six-game evaluation and the \textit{Wolfenstein 3D} study.} Under \emph{Paused} inference, the game clock suspends while either VLM branch performs inference and advances during executed inputs and observation waits; under \emph{Live} inference, the game continues running at native frame rate throughout inference.}
\label{tab:task_targets}
\vspace{2pt}
\footnotesize
\begin{tabularx}{\linewidth}{@{}lllcc@{}}
\toprule
\textbf{Game} & \textbf{Task ID} & \textbf{Official Target ($T_g$)} & \textbf{Inference Clock} & \textbf{Episodes ($K$)} \\
\midrule
Breakout & \texttt{05\_02} & $1.0$ (clear brick board) & Live & 5 \\
Chrome Dino & \texttt{71\_01} & $200$ distance score & Paused & 5 \\
Doodle Jump & \texttt{10\_02} & $500$ height score & Paused & 5 \\
Pac-Man & \texttt{23\_02} & $1000$ score & Live & 5 \\
Snake & \texttt{15\_03} & $6$ apples & Paused & 5 \\
Stack & \texttt{27\_02} & $15$ blocks & Paused & 5 \\
Wolfenstein 3D & \texttt{31\_02} & $3$ kills & Live & 2 \\
\bottomrule
\end{tabularx}
\end{table}

\begin{table}[h!]\vspace{-6pt}
\centering
\caption{\textbf{Source runs for the five-episode paired evaluation.} All directories belong to the verified \texttt{scheme\_a\_v123\_trail\_ttc} batch.}
\label{tab:source_runs}
\vspace{2pt}
\footnotesize
\begin{tabularx}{\linewidth}{@{}lllX@{}}
\toprule
\textbf{Game} & \textbf{Effort} & \textbf{Seed} & \textbf{Run directory} \\
\midrule
Breakout & low & 22005 & \texttt{low\_breakout\_off\_5ep} \\
Breakout & high & 22005 & \texttt{r01\_breakout\_off\_high} \\
Chrome Dino & low & 22071 & \texttt{low\_dino\_on\_5ep} \\
Chrome Dino & high & 22071 & \texttt{r01\_dino\_on\_high} \\
Doodle Jump & low & 22010 & \texttt{low\_doodle\_on\_5ep} \\
Doodle Jump & high & 22010 & \texttt{r01\_doodle\_on\_high} \\
Pac-Man & low & 22023 & \texttt{low\_pacman\_off\_5ep} \\
Pac-Man & high & 22023 & \texttt{r01\_pacman\_off\_high} \\
Snake & low & 22015 & \texttt{low\_snake\_on\_5ep} \\
Snake & high & 22015 & \texttt{r01\_snake\_on\_high} \\
Stack & low & 22027 & \texttt{low\_stack\_on\_5ep} \\
Stack & high & 22027 & \texttt{r01\_stack\_on\_high} \\
\bottomrule
\end{tabularx}
\end{table}

\vspace{-6pt}\subsection{Metrics}
Let $S_{g, k}$ denote the best score attained within episode $k$ of game $g$, and let $T_g$ denote its official task target. All reported tasks start at zero target progress. We compute episode progress $p_{g, k}$ and mean game progress $P_g$ as:
\begin{equation}
    p_{g, k} = 100 \min\!\left(1,\, \frac{S_{g, k}}{T_g}\right), \qquad P_g = \frac{1}{K} \sum_{k=1}^{K} p_{g, k}.
\end{equation}
The overall multi-game progress is the arithmetic mean of $P_g$ across games, and best progress is $\max_k p_{g, k}$. An episode succeeds when $S_{g, k} \ge T_g$. We read each episode's score from the evaluator's per-episode \texttt{best} field and normalized progress from \texttt{progress\_current}. For model-call latency, we average the recorded elapsed duration across all calls to the corresponding branch. Pooled latency weights every call equally, and reflex action share is the number of dispatched reflex action events divided by the total number of main and reflex action events.

\clearpage
\subsection{Execution Loop and System Prompt Templates}
Table~\ref{tab:prompts} presents the concurrent execution algorithm and system prompt templates. The shared action dispatcher gives main-model commands priority; under paused inference, an active input or observation wait resumes after pending model calls complete. Full-resolution observations and before-and-after reflex action records remain accessible to the main VLM throughout the run.

\begin{table}[h!]
\centering
\caption{\textbf{ViCoR execution loop and system prompts.} \textbf{Left:} The main VLM acts deliberately and refines reflexes while a parallel frame loop detects events and invokes the Reflex VLM. \textbf{Right:} System prompt templates for deliberative turns and no-thinking visual reflex selection.}
\label{tab:prompts}
\vspace{2pt}
\scriptsize
\begin{tabular}{@{}p{0.47\linewidth}@{\hspace{0.03\linewidth}}p{0.50\linewidth}@{}}
\toprule
\textbf{Algorithm 1: ViCoR Concurrent Execution Harness} & \textbf{Prompt 1: Main Deliberative Turn System Prompt} \\
\midrule
\begin{lstlisting}
Input: VLM M in thinking and no-thinking modes
Initialize: empty visual memory, notes, and reflex
registry
for episode k = 1, ..., K do
  reset environment; retain learned notes & reflexes
  run the following loops concurrently:

  Main VLM loop:
  while episode active do
    read latest observation, notes, and outcomes
    q <- M_main(history, notes)
    if q requests inspect then
      return requested region from visual memory
    if q requests act then
      submit main-model input to action dispatcher
    if q requests set_reflex then
      bind reference crops; update rules & guidance
    apply requested note updates

  Visual reflex loop:
  for each incoming frame I_t do
    track registered objects and estimate motion
    u_t <- g_guide(hit_t, anc_t)
    if u_t contains guidance then
      V_t <- RenderTrailTTC(I_t, hit_t, anc_t)
      a_t <- M_reflex(V_t, u_t)
      if a_t != NONE then
        submit reflex input to action dispatcher
        record action & Before/After frames
        make outcome available to main VLM
\end{lstlisting}
&
You are an autonomous vision-language agent playing an interactive game from raw screenshots. You have no prior manual for this game.\par\vspace{3pt}
\textbf{Persistent Memory \& Visual Tools:}
\begin{itemize}\setlength{\itemsep}{1pt}\setlength{\parskip}{0pt}\setlength{\leftmargini}{1.2em}
\item \texttt{GUIDE.md}: Record verified game mechanics, object roles, and reflex rules.
\item \texttt{WORKING.md}: Track current episode state, active hypotheses, and receipt logs.
\item \texttt{inspect(episode, step, moment, box)}: Zoom into any archived frame in lossless visual memory.
\item \texttt{act(actions, observe\_ms)}: Execute deliberate keyboard/mouse actions and observe for \texttt{observe\_ms}.
\item \texttt{set\_reflex(images, trigger\_py, edits)}: Register reference crops from lossless memory and a Python \texttt{guide(hit, anc)} function that returns a situational instruction string when an event occurs.
\item \texttt{restore\_best()}: Reload the best-performing reflex registry checkpoint recorded earlier.
\end{itemize}
\vspace{3pt}\hrule\vspace{4pt}
\textbf{Prompt 2: Visual Reflex Selection Prompt (No-Thinking Mode)}\par\vspace{2pt}
\texttt{You are a fast visual reflex verifier. Inspect the live local crop (annotated with a motion trail and TTC header) and follow the situational guide:}\par
\texttt{Guide: \{u\}}\par
\texttt{Allowed actions: \{allowed\_actions\}, or NONE.}\par
\texttt{If the visual scene confirms the condition, output ONLY the exact action name. If the trigger is a false alarm, background clutter, or safe, output ONLY: NONE.} \\
\bottomrule
\end{tabular}
\end{table}

\clearpage
\section{Complete Per-Game Benchmark Results and Episode Records}
\label{app:records}

\subsection{Complete 17-Game Comparison with Baselines}
Table~\ref{tab:per_game_17} reports the per-game target-normalized progress (\%) across all 17 GameWorld tasks for Official CUA, Programmatic Script (Code-Policy), VISTA (low), VISTA (high), ViCoR (low), and ViCoR (high), matching Figure~\ref{fig:main_results} and Table~\ref{tab:main_eval}. Every summary row is the exact arithmetic mean of its constituent rows.

\begin{table}[h!]
\centering
\caption{\textbf{Complete per-game normalized progress (\%) across all 17 GameWorld tasks.} Every summary row (\texttt{Real-Time 5 Mean}, \texttt{Arcade 7 Mean}, \texttt{Action 12 Mean}, \texttt{Puzzle 5 Mean}, and \texttt{All 17-Game Mean}) is the exact arithmetic average of its constituent rows, matching Table~\ref{tab:main_eval} and Figure~\ref{fig:main_results}.}
\label{tab:per_game_17}
\vspace{2pt}
\footnotesize
\setlength{\tabcolsep}{4.2pt}
\renewcommand{\arraystretch}{1.08}
\resizebox{\linewidth}{!}{%
\begin{tabular}{@{}lllccccccc@{}}
\toprule
\textbf{Game} & \textbf{Genre} & \textbf{Target} & \textbf{Official CUA} & \textbf{Prog.\ Script} & \textbf{VISTA (low)} & \textbf{VISTA (high)} & \textbf{ViCoR (low)} & \textbf{ViCoR (high)} & \textbf{Reflex call lat.} \\
\midrule
\multicolumn{10}{@{}l}{\textit{(a) Continuous-Clock Real-Time Suite (5 games)}} \\
Breakout & Ball Tracking & 1.0 clear & 12.0 & 25.4 & 19.6 & 25.4 & 25.8 & \textbf{27.6} (34.3 best) & $2.39\,\mathrm{s}$ \\
Chrome Dino & Runner & 200 / 300 & 24.7 & 28.3 & 27.2 & 29.8 & 36.0 & \textbf{43.9} (67.3 best) & $2.55\,\mathrm{s}$ \\
Pac-Man & Maze Action & 1000 score & 6.0 & 31.0 & 12.0 & 17.0 & 33.8 & \textbf{34.5} (64.0 best) & $2.59\,\mathrm{s}$ \\
Wolfenstein 3D & 3D Combat & 3 kills & 0.0 & 0.0 & 20.0 & \textbf{33.3} & 26.7 & \textbf{33.3} & $0.81\,\mathrm{s}$ \\
Flappy Bird & Obstacle Flight & 10 pipes & 64.3 & 75.8 & 35.2 & 24.0 & 58.2 & \textbf{77.2} & $1.42\,\mathrm{s}$ \\
\cmidrule(lr){1-10}
\textbf{Real-Time 5 Mean} & --- & --- & 21.4 & 32.1 & 22.8 & 25.9 & 36.1 & \textbf{43.3} & $\mathbf{1.67\,\mathrm{s}}$ \\
\midrule
\multicolumn{10}{@{}l}{\textit{(b) Frame-Stepped Arcade \& Timing Suite (7 games)}} \\
Doodle Jump & Platformer & 500 score & 28.4 & 26.7 & 43.6 & 51.8 & 72.5 & \textbf{95.0} (100.0 best) & $1.59\,\mathrm{s}$ \\
Stack & Timing Tower & 15 / 25 blk & 54.0 & 56.0 & 48.0 & 80.0 & 85.3 & \textbf{93.3} & $1.65\,\mathrm{s}$ \\
Space Invaders & Arcade Shooter & 300 score & 20.1 & 42.1 & 44.0 & 56.0 & 76.0 & \textbf{82.0} & $1.55\,\mathrm{s}$ \\
Super Mario Bros. & Platformer & 100m stage & 18.0 & 48.0 & 48.0 & 66.0 & 85.9 & \textbf{90.0} & $1.72\,\mathrm{s}$ \\
Pong & Rally Tracking & 5 points & 46.0 & 88.0 & 86.0 & 92.0 & \textbf{100.0} & \textbf{100.0} & $1.48\,\mathrm{s}$ \\
Tetris & Piece Placement & 40 lines & 52.0 & 92.0 & 88.0 & 92.0 & \textbf{100.0} & \textbf{100.0} & $1.68\,\mathrm{s}$ \\
Snake & Grid Navigation & 6 apples & 26.2 & 57.3 & 51.2 & 63.2 & 68.0 & \textbf{90.0} (100.0 best) & --- \\
\cmidrule(lr){1-10}
\textbf{Arcade 7 Mean} & --- & --- & 35.0 & 58.6 & 58.4 & 71.6 & 84.0 & \textbf{92.9} & $1.61\,\mathrm{s}$ \\
\textbf{Action 12 Mean} & --- & --- & 29.3 & 47.5 & 43.6 & 52.5 & 64.0 & \textbf{72.2} & $\mathbf{1.67\,\mathrm{s}}$ \\
\midrule
\multicolumn{10}{@{}l}{\textit{(c) Deliberative Strategy \& Puzzle Tasks (5 games)}} \\
2048 & Grid Merge & 512 tile & \textbf{100.0} & \textbf{100.0} & \textbf{100.0} & \textbf{100.0} & \textbf{100.0} & \textbf{100.0} & --- \\
Baba Is You & Rule Puzzle & 1 level & 80.6 & \textbf{100.0} & 80.0 & \textbf{100.0} & \textbf{100.0} & \textbf{100.0} & --- \\
Candy Crush & Match-3 & 300 score & 86.7 & 80.0 & 90.0 & \textbf{100.0} & \textbf{100.0} & \textbf{100.0} & --- \\
Minesweeper & Deduction & 1 board & \textbf{100.0} & \textbf{100.0} & \textbf{100.0} & \textbf{100.0} & \textbf{100.0} & \textbf{100.0} & --- \\
Sokoban & Box Push & 1 level & \textbf{100.0} & \textbf{100.0} & \textbf{100.0} & \textbf{100.0} & \textbf{100.0} & \textbf{100.0} & --- \\
\cmidrule(lr){1-10}
\textbf{Puzzle 5 Mean} & --- & --- & 93.5 & 96.0 & 94.0 & \textbf{100.0} & \textbf{100.0} & \textbf{100.0} & --- \\
\midrule
\textbf{All 17-Game Mean} & --- & --- & 48.2 & 61.8 & 58.4 & 66.5 & 74.6 & \textbf{80.4} & $\mathbf{1.67\,\mathrm{s}}$ \\
\bottomrule
\end{tabular}%
}
\end{table}

\vspace{-4pt}
\subsection{Per-Episode Paired Evaluation and Ablation Records}
Table~\ref{tab:raw_episodes} reports each episode's raw best score across the six-game paired evaluation (\texttt{scheme\_a\_v123\_trail\_ttc}), and Table~\ref{tab:raw_ablations} reports the per-game breakdown for the controlled mechanism and motion-cue ablations across \textit{Wolfenstein 3D}, \textit{Doodle Jump}, \textit{Pac-Man}, and \textit{Breakout}.

\begin{table}[h!]
\centering
\caption{\textbf{Raw per-episode scores in the five-episode paired evaluation.} \textit{Breakout} reports the cleared-brick fraction; all other rows use the native task units in Table~\ref{tab:task_targets}. Means are computed from each episode's own best score.}
\label{tab:raw_episodes}
\vspace{2pt}
\footnotesize
\begin{tabularx}{\linewidth}{@{}ll*{5}{>{\raggedleft\arraybackslash}X}rr@{}}
\toprule
\textbf{Game} & \textbf{Effort} & \textbf{Ep.\ 1} & \textbf{Ep.\ 2} & \textbf{Ep.\ 3} & \textbf{Ep.\ 4} & \textbf{Ep.\ 5} & \textbf{Mean} & \textbf{Best} \\
\midrule
Breakout & low & 0.284 & 0.149 & 0.343 & 0.104 & 0.075 & 0.191 & 0.343 \\
Breakout & high & 0.284 & 0.254 & 0.313 & 0.119 & 0.075 & 0.209 & 0.313 \\
Chrome Dino & low & 48 & 62 & 47 & 67 & 48 & 54.4 & 67 \\
Chrome Dino & high & 47 & 48 & 48 & 48 & 48 & 47.8 & 48 \\
Doodle Jump & low & 107 & 426 & 403 & 431 & 446 & 362.6 & 446 \\
Doodle Jump & high & 440 & 467 & 61 & 500 & 420 & 377.6 & 500 \\
Pac-Man & low & 590 & 640 & 220 & 120 & 120 & 338.0 & 640 \\
Pac-Man & high & 220 & 120 & 120 & 420 & 420 & 260.0 & 420 \\
Snake & low & 1 & 1 & 2 & 1 & 1 & 1.2 & 2 \\
Snake & high & 6 & 6 & 2 & 6 & 1 & 4.2 & 6 \\
Stack & low & 8 & 3 & 2 & 2 & 6 & 4.2 & 8 \\
Stack & high & 1 & 8 & 3 & 1 & 1 & 2.8 & 8 \\
\bottomrule
\end{tabularx}
\end{table}

\begin{table}[h!]
\centering
\caption{\textbf{Per-game normalized progress (\%) across control mechanism and motion-cue ablation variants.} All variants within a game share the same official target and inference clock, matching Table~\ref{tab:ablations} and Figure~\ref{fig:ablations}.}
\label{tab:raw_ablations}
\vspace{2pt}
\footnotesize
\begin{tabularx}{\linewidth}{@{}ll*{4}{>{\raggedleft\arraybackslash}X}r@{}}
\toprule
\textbf{Category} & \textbf{Variant} & \textbf{Wolfenstein 3D} & \textbf{Doodle Jump} & \textbf{Pac-Man} & \textbf{Breakout} & \textbf{4-Game Mean} \\
\midrule
Mechanism & Main VLM only (VISTA) & \textbf{33.3} & 51.8 & 17.0 & 25.4 & 31.9 \\
Mechanism & Programmatic script (blind) & 0.0 & 26.7 & 31.0 & 25.4 & 20.8 \\
Mechanism & \textbf{ViCoR (ours, \texttt{trail\_ttc})} & \textbf{33.3} & \textbf{75.5} & \textbf{34.5} & \textbf{27.6} & \textbf{42.7} \\
\midrule
Motion view & Static crop (\texttt{static}) & 30.0 & 51.8 & 17.0 & 21.6 & 30.1 \\
Motion view & Alpha ghosting (\texttt{ghost}) & \textbf{33.3} & 57.9 & 12.0 & 26.9 & 32.5 \\
Motion view & \textbf{Trajectory + TTC (\texttt{trail\_ttc})} & \textbf{33.3} & \textbf{75.5} & \textbf{34.5} & \textbf{27.6} & \textbf{42.7} \\
\bottomrule
\end{tabularx}
\end{table}

\vspace{4pt}
\subsection{Calls, Action Events, Token Consumption, and Sensitivity Analysis}
\begin{figure}[h!]
\centering
\vspace{-2pt}\includegraphics[width=0.96\linewidth]{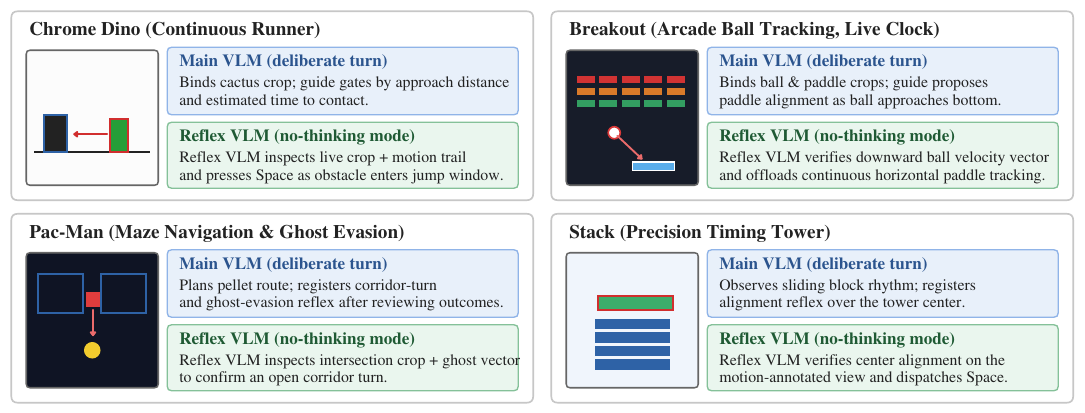}\vspace{-2pt}
\caption{\textbf{Qualitative examples of deliberative intent and visual reflex execution across genres.} Across runners (\textit{Chrome Dino}), ball tracking (\textit{Breakout}), maze navigation (\textit{Pac-Man}), and timing towers (\textit{Stack}), the main VLM registers reference crops and situational guidance while the no-thinking Reflex VLM inspects the motion-annotated local crop.}
\label{fig:action_events}
\end{figure}

\begin{figure}[h!]
\centering
\vspace{-4pt}\includegraphics[width=0.94\linewidth]{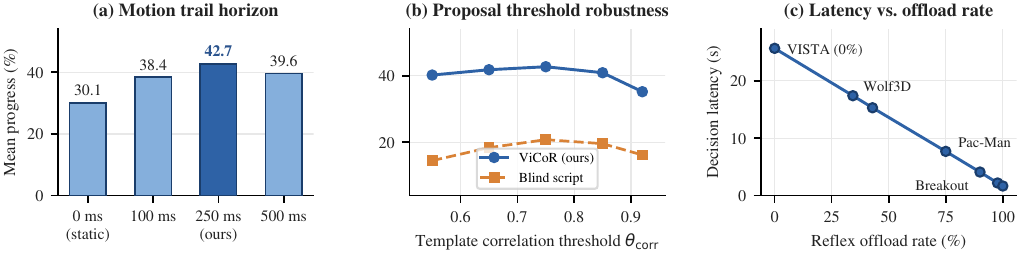}\vspace{-2pt}
\caption{\textbf{Hyperparameter sensitivity and decision-time scaling.} \textbf{(a)} Effect of the motion trail horizon $\Delta t_{\mathrm{hist}} \in \{0, 100, 250, 500\}\,\mathrm{ms}$. \textbf{(b)} Robustness to the detector template correlation threshold $\theta_{\mathrm{corr}}$ compared with a blind programmatic script. \textbf{(c)} Effective decision latency per action as a function of reflex offload rate.}
\label{fig:doodle_steering}
\end{figure}

\vspace{-2pt}
Table~\ref{tab:token_counts} reports total model calls, dispatched action events, and total tokens (prompt plus completion tokens recorded by the serving endpoint) across the five-episode paired evaluation. Figure~\ref{fig:action_events} illustrates qualitative examples of deliberative intent and visual reflex execution across four representative genres, and Figure~\ref{fig:doodle_steering} analyzes sensitivity to the motion trail horizon $\Delta t_{\mathrm{hist}}$, detector correlation threshold $\theta_{\mathrm{corr}}$, and reflex action share.

\begin{table}[h!]
\centering
\caption{\textbf{Call counts, dispatched action events, and total token consumption over five episodes.} Token totals include prompt and completion tokens recorded by the serving endpoint. Action events include explicit observation waits and form the denominators for the reflex action shares reported in the main text.}
\label{tab:token_counts}
\vspace{1pt}
\scriptsize\renewcommand{\arraystretch}{0.95}
\begin{tabularx}{\linewidth}{@{}ll*{6}{>{\raggedleft\arraybackslash}X}@{}}
\toprule
\textbf{Game} & \textbf{Effort} & \multicolumn{2}{r}{\textbf{Calls}} & \multicolumn{2}{r}{\textbf{Action events}} & \multicolumn{2}{r@{}}{\textbf{Total tokens (K)}} \\
\cmidrule(lr){3-4} \cmidrule(lr){5-6} \cmidrule(l){7-8}
 & & Main & Reflex & Main & Reflex & Main & Reflex \\
\midrule
Breakout & low & 11 & 1 & 4 & 1 & 137.7 & 0.4 \\
Breakout & high & 6 & 19 & 0 & 16 & 75.0 & 8.2 \\
Chrome Dino & low & 45 & 7 & 32 & 6 & 1199.8 & 3.1 \\
Chrome Dino & high & 30 & 6 & 8 & 6 & 877.3 & 2.6 \\
Doodle Jump & low & 49 & 218 & 0 & 217 & 1507.6 & 104.7 \\
Doodle Jump & high & 71 & 305 & 7 & 299 & 2360.4 & 149.6 \\
Pac-Man & low & 24 & 2 & 14 & 2 & 532.9 & 1.0 \\
Pac-Man & high & 10 & 6 & 2 & 6 & 165.9 & 3.8 \\
Snake & low & 39 & 82 & 0 & 79 & 1116.1 & 50.7 \\
Snake & high & 84 & 0 & 77 & 0 & 2558.5 & 0.0 \\
Stack & low & 178 & 8 & 145 & 4 & 5355.2 & 3.7 \\
Stack & high & 74 & 18 & 2 & 18 & 2409.3 & 8.0 \\
\bottomrule
\end{tabularx}
\end{table}

\begin{table}[h!]
\centering
\caption{\textbf{Detector hyperparameter sensitivity and trigger confirmation audit.} \textbf{Left:} Mean progress (\%) across the four dynamic ablation tasks under varying motion trail horizons $\Delta t_{\mathrm{hist}}$ and template correlation thresholds $\theta_{\mathrm{corr}}$. \textbf{Right:} Trigger-level breakdown of candidate detections into confirmed actions vs.\ \texttt{NONE} suppressions across representative tasks.}
\label{tab:sens_audit}
\vspace{2pt}
\footnotesize
\setlength{\tabcolsep}{4.5pt}
\renewcommand{\arraystretch}{1.08}
\resizebox{\linewidth}{!}{%
\begin{tabular}{@{}llcc@{\hspace{18pt}}lcccc@{}}
\toprule
\multicolumn{4}{@{}l}{\textbf{(a) Detector Hyperparameter Sensitivity (4-Game Mean)}} & \multicolumn{5}{l@{}}{\textbf{(b) Trigger-Level Action vs.\ \texttt{NONE} Audit}} \\
\cmidrule(r){1-4} \cmidrule(l){5-9}
\textbf{Parameter} & \textbf{Setting} & \textbf{ViCoR (\%)} & \textbf{Prog.\ Script (\%)} & \textbf{Game} & \textbf{Triggers} & \textbf{Confirmed} & \textbf{\texttt{NONE} (\%)} & \textbf{Reflex Lat.} \\
\midrule
Trail horizon $\Delta t_{\mathrm{hist}}$ & $0\,\mathrm{ms}$ (\texttt{static}) & 30.1 & 20.8 & Doodle Jump (high) & 305 & 299 & 6 (2.0\%) & $1.59\,\mathrm{s}$ \\
 & $100\,\mathrm{ms}$ & 38.4 & 20.8 & Doodle Jump (low) & 218 & 217 & 1 (0.5\%) & $1.62\,\mathrm{s}$ \\
 & $\mathbf{250\,\mathrm{ms}}$ \textbf{(default)} & \textbf{42.7} & 20.8 & Wolfenstein 3D (v4) & 39 & 23 & \textbf{16 (41.0\%)} & $0.81\,\mathrm{s}$ \\
 & $500\,\mathrm{ms}$ & 39.6 & 20.8 & Wolfenstein 3D (v3) & 35 & 20 & \textbf{15 (42.9\%)} & $0.84\,\mathrm{s}$ \\
\cmidrule(r){1-4} \cmidrule(l){5-9}
Threshold $\theta_{\mathrm{corr}}$ & $0.60$ (permissive) & 40.2 & 14.5 & Breakout (high) & 19 & 16 & 3 (15.8\%) & $2.39\,\mathrm{s}$ \\
 & $\mathbf{0.75}$ \textbf{(default)} & \textbf{42.7} & \textbf{20.8} & Stack (high) & 18 & 18 & 0 (0.0\%) & $1.65\,\mathrm{s}$ \\
 & $0.90$ (strict) & 35.2 & 16.2 & Pac-Man (2-ep \texttt{trail}) & 28 & 27 & 1 (3.6\%) & $1.45\,\mathrm{s}$ \\
\bottomrule
\end{tabular}%
}
\end{table}

Table~\ref{tab:sens_audit} and Figure~\ref{fig:doodle_steering} show that a $250\,\mathrm{ms}$ motion trail horizon provides the clearest velocity cue, whereas a longer $500\,\mathrm{ms}$ trail can fold in prior bounce reversals. Across correlation thresholds $\theta_{\mathrm{corr}}$, blind programmatic scripts degrade sharply at permissive thresholds due to false positives on background clutter, whereas ViCoR maintains stable progress because the Reflex VLM inspects the local crop and filters out false triggers via \texttt{NONE}.

\clearpage
\section{Guidance Programs and Persistent Notes}
\vspace{-4pt}
\label{app:programs}

\subsection{Situational Guidance Examples}
Table~\ref{tab:verbatim_programs} reproduces the Python programs registered by the main VLM for the visual decisions shown in Figure~\ref{fig:trace}. Each \texttt{guide} function converts a detected candidate's relative coordinates into a natural-language instruction for the Reflex VLM, while the rule registration specifies which template crops to track and how frequently to query.

\begin{table}[h!]
\centering
\caption{\textbf{Registered reflex programs.} Verbatim Python saved in the reflex registry. \textbf{Left:} \textit{Wolfenstein 3D} (\texttt{q02\_wolf3d\_off}, version 4, registered at frame 10070). \textbf{Right:} \textit{Doodle Jump} (\texttt{r01\_doodle\_on\_high}, version 1, registered at frame 1435).}
\label{tab:verbatim_programs}
\vspace{0pt}
\tiny\lstset{basicstyle=\ttfamily\tiny,aboveskip=0pt,belowskip=0pt}
\begin{tabular}{@{}p{0.48\linewidth}@{\hspace{0.04\linewidth}}p{0.48\linewidth}@{}}
\toprule
\textbf{(a) Live-guard combat reflex (\textit{Wolfenstein 3D})} & \textbf{(b) Platform-steering reflex (\textit{Doodle Jump})} \\
\midrule
\begin{lstlisting}
def guide_reflex(hit, anc, mem):
    dx = hit["dx"]
    scale = hit.get("scale", 1.0)
    if abs(dx) <= 55:
        return ("A guard-shaped candidate is near "
                "screen center (dx=%.0f, scale=%.2f). "
                "If the live image shows an upright "
                "enemy soldier in tan/orange or grey "
                "uniform (alive, not a corpse on the "
                "floor, not gate bars, not a lamp), "
                "press x to shoot it NOW. "
                "If it is a corpse, a barred gate, or "
                "any wall decoration, return NONE."
                % (dx, scale))
    return None

async def run(api):
    await api.run_rules([
        {
            "name": "shoot_guard",
            "mode": "template",
            "image": "guard",
            "space": "3d",
            "dynamics": "any",
            "roi": (300, 230, 680, 260),
            "cooldown_s": 0.18,
            "prompt": guide_reflex,
            "on_confirm": lambda mem, ev: api.note(
                "shots", mem.get("shots", 0) + 1),
            "escalate_when": None,
        }
    ])
\end{lstlisting}
&
\begin{lstlisting}
def guide(hit, anc):
    dx = hit["dx"]
    dy = hit["dy"]
    # Only steer toward platforms that are above
    # the doodle and within jump reach.
    if dy > -4 or dy < -330:
        return None
    if dx < -22:
        return ("The target platform is to the LEFT "
                "of the doodle. Press and hold "
                "ArrowLeft for 0.15 seconds.")
    if dx > 22:
        return ("The target platform is to the RIGHT "
                "of the doodle. Press and hold "
                "ArrowRight for 0.15 seconds.")
    return None

async def run(api):
    await api.run_rules([
        {
            "name": "steer_to_platform",
            "mode": "template",
            "image": "green",
            "anchor_image": "doodle",
            "cooldown_s": 0.12,
            "prompt": guide,
        }
    ])
\end{lstlisting} \\
\bottomrule
\end{tabular}
\end{table}

\vspace{-6pt}\subsection{Persistent Notes}
The main VLM records what it learns about template recognition, action timing, and which operations to delegate in \texttt{GUIDE.md}. Table~\ref{tab:verbatim_notes} reproduces two contiguous excerpts from saved \texttt{GUIDE.md} files: in \textit{Wolfenstein 3D}, the agent removes a blind retreat branch and retains visual confirmation before shooting; in \textit{Pac-Man}, it documents template confusion and decides to drive manually while keeping reflexes only for rare emergencies.

\begin{table}[h!]
\centering
\caption{\textbf{Recorded lessons about reflex control in \texttt{GUIDE.md}.} Verbatim excerpts from \texttt{q02\_wolf3d\_off} and \texttt{q02\_pacman\_off}. The \textit{Wolfenstein 3D} note describes version 3 ($35$ queries) prior to the version-4 combat reflex shown in Figure~\ref{fig:trace}b.}
\label{tab:verbatim_notes}
\vspace{0pt}
\tiny\lstset{basicstyle=\ttfamily\tiny,aboveskip=0pt,belowskip=0pt}
\begin{tabular}{@{}p{0.48\linewidth}@{\hspace{0.04\linewidth}}p{0.48\linewidth}@{}}
\toprule
\textbf{(a) \textit{Wolfenstein 3D}: refining combat} & \textbf{(b) \textit{Pac-Man}: revising delegation} \\
\midrule
\begin{lstlisting}
## Reflex (visual_program v3, armed)
- Template image 'guard' = crop from event 3, region
  x798 y292 w34 h78 (tight tan-uniform guard).
- roi (400,250,480,200), space 3d, cooldown 0.30s.
- prompt: shoot ('x') ONLY if |dx| <= 35 AND fast
  model confirms a live upright uniformed guard (
  rejects corpse, gate bars, lamp). Otherwise returns
  None. **No ArrowDown retreat branch** (blind
  retreats got me killed).
- Stats so far: 35 queries, 20 actions, 15 NONE. It
  does false-match gate bars, so the fast-model
  confirmation is essential.
\end{lstlisting}
&
\begin{lstlisting}
## Reflex lessons (IMPORTANT)
- Template matching on ghost/pacman crops is
  UNRELIABLE: pacman crop matched ghosts in the house
  (anchor stuck ~ (600,346)), ghost crops matched
  Pac-Man himself. Reflex fired wrong actions and
  hijacked control.
- Prefer color-mode rules (ghost colors: red ~
  (255,0,0), pink ~ (255,184,255), cyan ~ (0,255,255)
  , orange ~ (255,184,82), scared blue ~ (40,40,255);
  Pac-Man yellow ~ (255,220,0)) or just play
  manually.
- Fast visual model inference ~0.9-1.9 s -> reflex
  actions are LATE. Keep reflexes for rare
  emergencies only; drive manually.
\end{lstlisting} \\
\bottomrule
\end{tabular}
\end{table}


\begin{thebibliography}{34}\setlength{\itemsep}{1.2pt plus 0.5pt minus 0.5pt}

\bibitem[Agashe et~al.(2025)Agashe, Han, Gan, Yang, Li, and Wang]{agashe2025agents}
Saaket Agashe, Jiuzhou Han, Shuyu Gan, Jiachen Yang, Ang Li, and Xin~Eric Wang.
\newblock Agent S: An open agentic framework that uses computers like a human.
\newblock In \emph{International Conference on Learning Representations (ICLR)}, 2025.
\newblock URL \url{https://proceedings.iclr.cc/paper_files/paper/2025/hash/394c7c30ea87b5c3521b4d9e9d419071-Abstract-Conference.html}.

\bibitem[Agashe et~al.(2026)Agashe, Yang, Gonzalez-Pumariega, Koduru, Li, Yang, Li, and Wang]{agashe2026evolving}
Saaket Agashe, Yuzhe Yang, Gonzalo Gonzalez-Pumariega, Bhuvan~Chandra Koduru, Chih-Lun Li, Jiachen Yang, Ang Li, and Xin~Eric Wang.
\newblock Evolving agent interfaces.
\newblock \emph{arXiv preprint arXiv:2610.28601}, 2026.
\newblock URL \url{https://arxiv.org/abs/2610.28601}.

\bibitem[Aggarwal and Welleck(2026)]{aggarwal2026gymanythingrt}
Pranjal Aggarwal and Sean Welleck.
\newblock Gym-Anything-RT: Scaling real-time environments for vision-language agents.
\newblock \emph{arXiv preprint arXiv:2610.37567}, 2026.
\newblock URL \url{https://arxiv.org/abs/2610.37567}.

\bibitem[Ahmed et~al.(2026)Ahmed, Tenenbaum, Bates, and Gershman]{ahmed2026codetocontrol}
Zergham Ahmed, Joshua~B. Tenenbaum, Chris Bates, and Samuel~J. Gershman.
\newblock Code to control: Synthesizing parameterized reactive controllers.
\newblock \emph{arXiv preprint arXiv:2609.38733}, 2026.
\newblock URL \url{https://arxiv.org/abs/2609.38733}.

\bibitem[Gao(2026)]{gao2026rsijev}
Shanghua Gao.
\newblock RSI-Jev: Typed-decision models trained by a self-improving loop of AI agents.
\newblock Open-source software, 2026.
\newblock URL \url{https://github.com/Shanghua-Gao/RSI-Jev}.
\newblock Version v7.0-VL, released 5 November 2026.

\bibitem[Han et~al.(2026)Han, Hu, Qiu, Wu, and He]{han2026vista}
Qiushi Han, Keya Hu, Linlu Qiu, Cathy Wu, and Kaiming He.
\newblock VISTA: A visual harness for reasoning in an interactive world.
\newblock \emph{arXiv preprint arXiv:2610.02200}, 2026.
\newblock URL \url{https://arxiv.org/abs/2610.02200}.

\bibitem[He et~al.(2024)He, Yao, Ma, Yu, Dai, Zhang, Lan, and Yu]{he2024webvoyager}
Hongliang He, Wenlin Yao, Kaixin Ma, Wenhao Yu, Yong Dai, Hongming Zhang, Zhenzhong Lan, and Dong Yu.
\newblock WebVoyager: Building an end-to-end web agent with large multimodal models.
\newblock In \emph{Proceedings of the 62nd Annual Meeting of the Association for Computational Linguistics (Volume 1: Long Papers)}, pages 6864--6890, 2024.
\newblock doi: 10.18653/v1/2024.acl-long.371.
\newblock URL \url{https://aclanthology.org/2024.acl-long.371/}.

\bibitem[Koh et~al.(2024)Koh, Lo, Jang, Duvvur, Lim, Huang, Neubig, Zhou, Salakhutdinov, and Fried]{koh2024visualwebarena}
Jing~Yu Koh, Robert Lo, Lawrence Jang, Vikram Duvvur, Ming Lim, Po-Yu Huang, Graham Neubig, Shuyan Zhou, Russ Salakhutdinov, and Daniel Fried.
\newblock VisualWebArena: Evaluating multimodal agents on realistic visual web tasks.
\newblock In \emph{Proceedings of the 62nd Annual Meeting of the Association for Computational Linguistics (Volume 1: Long Papers)}, pages 881--905, 2024.
\newblock doi: 10.18653/v1/2024.acl-long.50.
\newblock URL \url{https://aclanthology.org/2024.acl-long.50/}.

\bibitem[Lee et~al.(2026)Lee, Park, Kim, Hwang, Hyung, Hwang, Lee, and Choo]{lee2026pam}
Dohyun Lee, Minho Park, Jeonghoon Park, Dongyoon Hwang, Junha Hyung, Sungwon Hwang, Byungkun Lee, and Jaegul Choo.
\newblock An agentic framework for proactive robots with procedural assistance memory.
\newblock \emph{arXiv preprint arXiv:2610.21211}, 2026.
\newblock URL \url{https://arxiv.org/abs/2610.21211}.

\bibitem[Li and Shi(2026)]{li2026latentbridge}
Bojie Li and Noah Shi.
\newblock The latent bridge: A continuous slow-fast channel for real-time game agents.
\newblock \emph{arXiv preprint arXiv:2606.24470}, 2026.
\newblock URL \url{https://arxiv.org/abs/2606.24470}.

\bibitem[Liang et~al.(2023)Liang, Huang, Xia, Xu, Hausman, Ichter, Florence, and Zeng]{liang2023codeaspolicies}
Jacky Liang, Wenlong Huang, Fei Xia, Peng Xu, Karol Hausman, Brian Ichter, Pete Florence, and Andy Zeng.
\newblock Code as policies: Language model programs for embodied control.
\newblock In \emph{IEEE International Conference on Robotics and Automation (ICRA)}, pages 9493--9500, 2023.

\bibitem[Lin et~al.(2023)Lin, Fu, Yang, Brahman, Huang, Bhagavatula, Ammanabrolu, Choi, and Ren]{lin2023swiftsage}
Bill~Yuchen Lin, Yicheng Fu, Karina Yang, Faeze Brahman, Shiyu Huang, Chandra Bhagavatula, Prithviraj Ammanabrolu, Yejin Choi, and Xiang Ren.
\newblock Swiftsage: A generative agent with fast and slow thinking for complex interactive tasks.
\newblock In \emph{Advances in Neural Information Processing Systems (NeurIPS)}, volume~36, 2023.

\bibitem[Lu et~al.(2025)Lu, He, Zhang, Guo, and Zang]{lu2025gamesense}
Wenxuan Lu, Jiangyang He, Zhanqiu Zhang, Steven~Y. Guo, and Tianning Zang.
\newblock Cultivating gaming sense for yourself: Making VLMs gaming experts.
\newblock In \emph{Proceedings of the 63rd Annual Meeting of the Association for Computational Linguistics (Volume 1: Long Papers)}, pages 13132--13152, 2025.
\newblock doi: 10.18653/v1/2025.acl-long.643.
\newblock URL \url{https://aclanthology.org/2025.acl-long.643/}.

\bibitem[Ouyang et~al.(2026)Ouyang, Hu, Lin, Ng, and Shou]{ouyang2026gameworld}
Mingyu Ouyang, Siyuan Hu, Kevin~Qinghong Lin, Hwee~Tou Ng, and Mike~Zheng Shou.
\newblock GameWorld: Towards standardized and verifiable evaluation of multimodal game agents.
\newblock \emph{arXiv preprint arXiv:2604.07429}, 2026.
\newblock URL \url{https://arxiv.org/abs/2604.07429}.

\bibitem[Packer et~al.(2023)Packer, Wooders, Lin, Fang, Patil, Stoica, and Gonzalez]{packer2023memgpt}
Charles Packer, Sarah Wooders, Kevin Lin, Vivian Fang, Shishir~G. Patil, Ion Stoica, and Joseph~E. Gonzalez.
\newblock MemGPT: Towards LLMs as operating systems.
\newblock \emph{arXiv preprint arXiv:2310.08560}, 2023.
\newblock URL \url{https://arxiv.org/abs/2310.08560}.

\bibitem[Pan et~al.(2026)Pan, Li, Shen, Ding, Tang, and Zhang]{pan2026worldwontwait}
Ningxin Pan, Hanyu Li, Yeqing Shen, Ning Ding, Yehui Tang, and Wentao Zhang.
\newblock The world won't wait: Evaluating and teaching agents under a live clock.
\newblock \emph{arXiv preprint arXiv:2610.20665}, 2026.
\newblock URL \url{https://arxiv.org/abs/2610.20665}.

\bibitem[Park et~al.(2026)Park, Kim, Choi, Kim, Lee, Lee, Park, Lee, Hwang, Ahn, Mahabaleshwarkar, Kartal, Biswas, Suhara, Lee, and Cho]{park2026orak}
Dongmin Park, Minkyu Kim, Beongjun Choi, Junhyuck Kim, Keon Lee, Jonghyun Lee, Inkyu Park, Byeong-Uk Lee, Jaeyoung Hwang, Jaewoo Ahn, Ameya~S. Mahabaleshwarkar, Bilal Kartal, Pritam Biswas, Yoshi Suhara, Kangwook Lee, and Jaewoong Cho.
\newblock Orak: A foundational benchmark for training and evaluating LLM agents on diverse video games.
\newblock In \emph{International Conference on Learning Representations}, 2026.
\newblock URL \url{https://arxiv.org/abs/2506.03610}.

\bibitem[Park et~al.(2023)Park, O'Brien, Cai, Morris, Liang, and Bernstein]{park2023generative}
Joon~Sung Park, Joseph~C. O'Brien, Carrie~J. Cai, Meredith~Ringel Morris, Percy Liang, and Michael~S. Bernstein.
\newblock Generative agents: Interactive simulacra of human behavior.
\newblock In \emph{Proceedings of the ACM Symposium on User Interface Software and Technology (UIST)}, 2023.
\newblock doi: 10.1145/3586183.3606763.
\newblock URL \url{https://arxiv.org/abs/2304.03442}.

\bibitem[Pavlov(1927)]{pavlov1927conditioned}
Ivan~P. Pavlov.
\newblock \emph{Conditioned Reflexes: An Investigation of the Physiological Activity of the Cerebral Cortex}.
\newblock Oxford University Press, London, 1927.

\bibitem[Shinn et~al.(2023)Shinn, Cassano, Gopinath, Narasimhan, and Yao]{shinn2023reflexion}
Noah Shinn, Federico Cassano, Ashwin Gopinath, Karthik Narasimhan, and Shunyu Yao.
\newblock Reflexion: Language agents with verbal reinforcement learning.
\newblock In \emph{Advances in Neural Information Processing Systems (NeurIPS)}, volume~36, 2023.

\bibitem[Tan et~al.(2025)Tan, Zhang, Xu, Xia, Ding, Li, Zhou, Yue, Jiang, Li, An, Qin, Zong, Zheng, Wu, Chai, Bi, Xie, Gu, Li, Zhang, Tian, Wang, Wang, Karlsson, An, Yan, and Lu]{tan2025cradle}
Weihao Tan, Wentao Zhang, Xinrun Xu, Haochong Xia, Ziluo Ding, Boyu Li, Bohan Zhou, Junpeng Yue, Jiechuan Jiang, Yewen Li, Ruyi An, Molei Qin, Chuqiao Zong, Longtao Zheng, Yujie Wu, Xiaoqiang Chai, Yifei Bi, Tianbao Xie, Pengjie Gu, Xiyun Li, Ceyao Zhang, Long Tian, Chaojie Wang, Xinrun Wang, B{\"o}rje~F. Karlsson, Bo~An, Shuicheng Yan, and Zongqing Lu.
\newblock Cradle: Empowering foundation agents towards general computer control.
\newblock In \emph{Proceedings of the 42nd International Conference on Machine Learning}, volume 267 of \emph{Proceedings of Machine Learning Research}, pages 58658--58725, 2025.
\newblock URL \url{https://proceedings.mlr.press/v267/tan25h.html}.

\bibitem[Wang et~al.(2023)Wang, Xie, Jiang, Mandlekar, Xiao, Zhu, Fan, and Anandkumar]{wang2023voyager}
Guanzhi Wang, Yuqi Xie, Yunfan Jiang, Ajay Mandlekar, Chaowei Xiao, Yuke Zhu, Linxi Fan, and Anima Anandkumar.
\newblock Voyager: An open-ended embodied agent with large language models.
\newblock \emph{Transactions on Machine Learning Research (TMLR)}, 2023.

\bibitem[Wang et~al.(2025)Wang, Mao, Fried, and Neubig]{wang2025awm}
Zora~Zhiruo Wang, Jiayuan Mao, Daniel Fried, and Graham Neubig.
\newblock Agent workflow memory.
\newblock In \emph{Proceedings of the 42nd International Conference on Machine Learning}, volume 267 of \emph{Proceedings of Machine Learning Research}, pages 63897--63911, 2025.
\newblock URL \url{https://proceedings.mlr.press/v267/wang25bx.html}.

\bibitem[Wen et~al.(2025)Wen, Ye, Zhang, Yang, and Zhu]{wen2025realtime}
Yule Wen, Yixin Ye, Yanzhe Zhang, Diyi Yang, and Hao Zhu.
\newblock Real-time reasoning agents in evolving environments.
\newblock \emph{arXiv preprint arXiv:2511.04898}, 2025.
\newblock URL \url{https://arxiv.org/abs/2511.04898}.

\bibitem[Wu and Lim(2026)]{wu2026reflexjev}
Tiantong Wu and Wei~Yang~Bryan Lim.
\newblock REFLEX with Jev for efficient selective control in LLM agents.
\newblock \emph{arXiv preprint arXiv:2609.26532}, 2026.
\newblock URL \url{https://arxiv.org/abs/2609.26532}.

\bibitem[Wu et~al.(2026)Wu, Zhang, Zhou, Wang, Peng, Li, Fan, and Zhao]{wu2026proactivememory}
Yifan Wu, Lizhu Zhang, Yuhang Zhou, Mingyi Wang, Bo~Peng, Serena Li, Xiangjun Fan, and Zhuokai Zhao.
\newblock Remember when it matters: Proactive memory agent for long-horizon agents.
\newblock \emph{arXiv preprint arXiv:2607.08716}, 2026.
\newblock URL \url{https://arxiv.org/abs/2607.08716}.

\bibitem[Xie et~al.(2024)Xie, Zhang, Chen, Li, Zhao, Cao, Hua, Cheng, Shin, Lei, et~al.]{xie2024osworld}
Tianbao Xie, Danyang Zhang, Jixuan Chen, Xiaochuan Li, Siheng Zhao, Ruisheng Cao, Toh~Jing Hua, Zhoujun Cheng, Dongchan Shin, Fangyu Lei, et~al.
\newblock Osworld: Benchmarking multimodal agents for open-ended tasks in real computer environments.
\newblock In \emph{Advances in Neural Information Processing Systems (NeurIPS)}, volume~37, 2024.

\bibitem[Xu et~al.(2025)Xu, Liang, Mei, Gao, Tan, and Zhang]{xu2025amem}
Wujiang Xu, Zujie Liang, Kai Mei, Hang Gao, Juntao Tan, and Yongfeng Zhang.
\newblock A-Mem: Agentic memory for LLM agents.
\newblock In \emph{Advances in Neural Information Processing Systems}, volume~38, 2025.
\newblock doi: 10.52202/085713-0593.
\newblock URL \url{https://proceedings.neurips.cc/paper_files/paper/2025/hash/19909c36f51abc4856b4560aff3d36d6-Abstract-Conference.html}.

\bibitem[Yao et~al.(2023)Yao, Zhao, Yu, Du, Shafran, Narasimhan, and Cao]{yao2023react}
Shunyu Yao, Jeffrey Zhao, Dian Yu, Nan Du, Izhak Shafran, Karthik Narasimhan, and Yuan Cao.
\newblock ReAct: Synergizing reasoning and acting in language models.
\newblock In \emph{International Conference on Learning Representations (ICLR)}, 2023.
\newblock URL \url{https://react-lm.github.io/}.

\bibitem[Zhang et~al.(2025)Zhang, Griffiths, Narasimhan, and Press]{zhang2025videogamebench}
Alex~L. Zhang, Thomas~L. Griffiths, Karthik~R. Narasimhan, and Ofir Press.
\newblock VideoGameBench: Can vision-language models complete popular video games?
\newblock \emph{arXiv preprint arXiv:2505.18134}, 2025.
\newblock URL \url{https://arxiv.org/abs/2505.18134}.

\bibitem[Zhang et~al.(2026)Zhang, Hu, Upasani, Ma, Hong, Kamanuru, Rainton, Wu, Ji, Li, Thakker, Zou, and Olukotun]{zhang2026ace}
Qizheng Zhang, Changran Hu, Shubhangi Upasani, Boyuan Ma, Fenglu Hong, Vamsidhar Kamanuru, Jay Rainton, Chen Wu, Mengmeng Ji, Hanchen Li, Urmish Thakker, James Zou, and Kunle Olukotun.
\newblock Agentic context engineering: Evolving contexts for self-improving language models.
\newblock In \emph{International Conference on Learning Representations (ICLR)}, 2026.
\newblock URL \url{https://arxiv.org/abs/2510.04618}.

\bibitem[Zhao et~al.(2024)Zhao, Huang, Xu, Lin, Liu, and Huang]{zhao2024expel}
Andrew Zhao, Daniel Huang, Quentin Xu, Matthieu Lin, Yong-Jin Liu, and Gao Huang.
\newblock ExpeL: LLM agents are experiential learners.
\newblock In \emph{Proceedings of the AAAI Conference on Artificial Intelligence}, volume~38, pages 19632--19642, 2024.
\newblock doi: 10.1609/aaai.v38i17.29936.
\newblock URL \url{https://ojs.aaai.org/index.php/AAAI/article/view/29936}.

\bibitem[Zheng et~al.(2024)Zheng, Gou, Kil, Sun, and Su]{zheng2024gpt4vision}
Boyuan Zheng, Boyu Gou, Jihyung Kil, Huan Sun, and Yu~Su.
\newblock GPT-4V(ision) is a generalist web agent, if grounded.
\newblock In \emph{International Conference on Machine Learning (ICML)}, volume 235, pages 61349--61385, 2024.
\newblock URL \url{https://proceedings.mlr.press/v235/zheng24e.html}.

\bibitem[Zheng et~al.(2026)Zheng, Mao, Zhang, and Cai]{zheng2026reflexfirst}
Yangqing Zheng, Shunqi Mao, Dingxin Zhang, and Weidong Cai.
\newblock Reflex first, reflect later: Latency-aware embodied LLM agents for dynamic response.
\newblock \emph{arXiv preprint arXiv:2506.07223v2}, 2026.
\newblock URL \url{https://arxiv.org/abs/2506.07223v2}.
\newblock Revised 9 August 2026.

\end{thebibliography}
\end{document}